%% file: arxiv.tex
\pdfoutput=1

\documentclass[11pt]{article}

\usepackage[]{acl}

\usepackage{times}
\usepackage{latexsym}
\usepackage{amsfonts}
\usepackage{arydshln}

\usepackage[T1]{fontenc}

\usepackage[utf8]{inputenc}

\usepackage{microtype}
\usepackage{comment}
\usepackage[ruled,vlined]{algorithm2e}
\usepackage{amsmath}
\usepackage{inconsolata}
\usepackage{multirow}
\usepackage{makecell}
\usepackage{graphicx}
\usepackage{booktabs}
\usepackage{subfig}
\usepackage{pifont}
\usepackage{enumitem}
\usepackage{algpseudocode}
\usepackage{soul}
\usepackage[most]{tcolorbox}
\tcbuselibrary{listings}
\tcbuselibrary{skins}
\tcbuselibrary{breakable}
\usepackage{bbding}
\usepackage{scalerel}
\usepackage{colortbl}
\definecolor{mintleaf}{RGB}{0, 184, 148}
\definecolor{dm-green-600}{RGB}{46,125,50}
\definecolor{dm-blue-500}{RGB}{0, 69, 177}
\definecolor{dm-purple-500}{RGB}{105,50,230}
\definecolor{mysilver}{RGB}{128,129,128}
\definecolor{my_green}{RGB}{0, 176, 80}
\definecolor{my_yellow}{RGB}{255,165,0}
\definecolor{my_red}{RGB}{255, 0, 0}
\definecolor{my_purple}{RGB}{126, 100, 158}
\definecolor{my_blue}{RGB}{49, 133, 155}
\definecolor{case_purple}{RGB}{160, 43, 147}
\definecolor{case_blue}{RGB}{15, 158, 213}

\title{SA\textsc{n}D: Boosting LLM Agents with Self-Taught Action Deliberation}

\author{Yu Xia$^{1}$ \quad Yiran Shen$^{1}$ \quad Junda Wu$^{1}$ \quad Tong Yu$^{2}$ \quad Sungchul Kim$^{2}$ \\ {\bf Ryan A. Rossi$^{2}$ \quad\; Lina Yao$^{3,4}$ \quad\; Julian McAuley$^{1}$} \\ 
$^{1}$University of California San Diego \qquad 
$^{2}$Adobe Research\\
$^{3}$University of New South Wales \quad 
$^{4}$CSIRO’s Data61\\
\texttt{\{yux078, jes038, juw069, jmcauley\}@ucsd.edu} \\ \texttt{\{tyu, sukim, ryrossi\}@adobe.com}\quad \texttt{lina.yao@unsw.edu.au} }

\begin{document}
\maketitle
\input{0-abstract}

\input{1-intro}

\input{2-related_work}

\input{3-methodology}

\input{4-experiment}

\input{5-result}

\input{6-conclusion}

\bibliography{anthology, custom}

\appendix

\input{7-appendix}

\end{document}

%% file: 0-abstract.tex
\begin{abstract}

Large Language Model (LLM) agents are commonly tuned with supervised finetuning on ReAct-style expert trajectories or preference optimization over pairwise rollouts. 
Most of these methods focus on imitating specific expert behaviors or promoting chosen reasoning thoughts and actions over rejected ones.
However, without reasoning and comparing over alternative actions, LLM agents finetuned with these methods may over-commit towards seemingly plausible but suboptimal actions due to limited action space exploration.
To address this, in this paper we propose Self-taught Actio\textsc{n} Deliberation (SA\textsc{n}D) framework, enabling LLM agents to explicitly deliberate over candidate actions before committing to one. 
To tackle the challenges of when and what to deliberate given large action space and step-level action evaluation, we incorporate self-consistency action sampling and execution-guided action critique to help synthesize step-wise action deliberation thoughts using the base model of the LLM agent.
In an iterative manner, the deliberation trajectories are then used to finetune the LLM agent itself.
Evaluating on two representative interactive agent tasks, SA\textsc{n}D achieves an average 20\% improvement over supervised finetuning on initial expert data and also outperforms state-of-the-art agent tuning approaches.

\end{abstract}

%% file: 1-intro.tex
\section{Introduction}

Large language models (LLMs) have recently been cast as agents that read instructions, reason through intermediate thoughts, and execute actions interacting with external environments such as web navigation \cite{nakano2021webgpt, yao2022webshop, nguyen-etal-2025-gui}, embodied household tasks \cite{shridhar2020alfworld}, or scientific experiments \cite{wang2022scienceworld}.
Early prompting-based methods such as ReAct \cite{yao2023react, wu-etal-2025-doc, wang2025dice} interleave chain-of-thoughts and actions, enabling the LLM to plan and gather new information in context.
To obtain more reliable LLM agents, recent works apply supervised finetuning on expert ReAct-style trajectories \cite{chen2023fireact, zeng-etal-2024-agenttuning, chen2024agent, wang-etal-2025-nat, chen2025atlas}, or directly optimize on agent trajectory preference pairs \cite{song-etal-2024-trial, xiong-etal-2024-watch, shi-etal-2024-direct}.

\begin{figure}[!t]
    \centering
    {\includegraphics[width=0.48\textwidth]{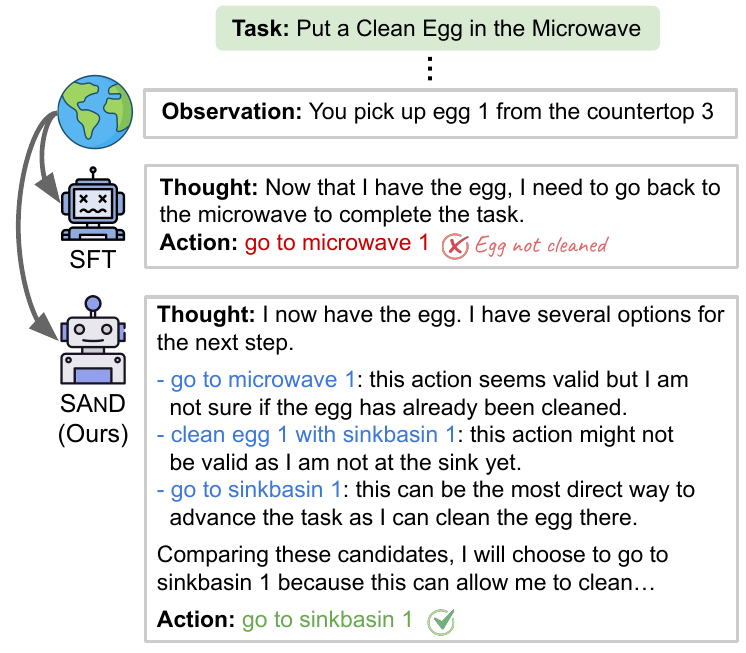}}
    \caption{An illustrative example of an LLM agent task, where SFT trained agent \cite{zeng-etal-2024-agenttuning} over-commits to an seemingly plausible but suboptimal action while our SA\textsc{n}D tuned agent learns to deliberate over candidate actions before choosing the best action.}
    \label{fig:intro} 
    \vspace{-1em}
\end{figure}

Although effective, these approaches imitate expert actions or simply rank chosen actions over rejected actions and expose the model to mostly the reference action and corresponding rationale at each decision point.
Without effectively exploring the action space, the agent seldom learns explicitly why the chosen action wins over plausible alternatives.
As a result, the finetuned LLM agent can over-commit to superficially reasonable yet suboptimal actions, a failure mode also observed in self-consistency studies of LLMs \cite{wang2023selfconsistency, xia-etal-2024-hallucination, liang2024internal}.
Such behavior also hurts the generalization performance of LLM agents to unfamiliar scenarios.

To address this, in this paper we aim to teach LLM agent to deliberate by first generating several candidate actions for the current state, evaluating and comparing their likely outcomes, and then commit only after this evaluation.
We propose Self-taught Actio\textsc{n} Deliberation (SA\textsc{n}D) framework to instantiate this idea by teaching the LLM agent with the deliberation thoughts synthesized by the base version of itself.
However, as the action space of LLM agent tasks is often large or even unbounded \cite{yao2022webshop, lin2025qlass}, it is intractable to deliberate over all actions and also inefficient to deliberate at every single step.
To further tackle the challenge of when and what to deliberate, we devise self-consistency action sampling along expert trajectories to sample uncertain candidate actions of LLM agent at non-trivial decision making steps.
To provide more informative and grounded step-level evaluations for each sampled candidate action, we utilize executed rollouts of each action to guide the critique generation. 
The action critiques are utilized to synthesize an action deliberation thought using the base LLM, which augments the initial expert trajectory and constructs deliberation trajectories for iterative finetuning of the LLM agent.
Experiments on two interactive tasks demonstrate the advantage of our methods compared with strong agent tuning baselines.
In summary, we make the following contributions:

\begin{itemize}[left=0pt, parsep=0pt, partopsep=0pt]
    \item To teach LLM agents better explore the action space, we propose Self-taught Actio\textsc{n} Deliberation (SA\textsc{n}D), a self-learning framework teaching LLM agents to deliberatively reason over candidate actions before choosing one.
    \item To tackle the challenge of when and what to deliberate given large action space and step-level action evaluation, we devise self-consistency action sampling and execution-guided action critique to help synthesize high-quality deliberative reasoning thoughts for iterative finetuning.
    \item Experiments on two representative interactive agent tasks demonstrate the advantage of our method with an average 20\% improvement over supervised finetuning on initial expert data and outperforming strong agent tuning baselines.
\end{itemize}

%% file: 2-related_work.tex
\section{Related Work}

\subsection{LLM Agents Tuning}
Recent efforts in tuning LLM agents have progressed from failure recovery heuristics towards more structured policy refinement. 
Early work such as FiReAct \cite{chen2023fireact} showed that adding explicit failure-reflection demonstrations improves LLM agent robustness. AgentTuning \cite{zeng-etal-2024-agenttuning} uses high-quality trajectories to finetune an instruction model for multi-turn interactions.
ETO \cite{song-etal-2024-trial} retains exploratory trajectories and contrasts them with expert trajectories for agent optimization, 
while IPR \cite{xiong-etal-2024-watch} obtain step-level rewards for iterative preference refinement. 
DMPO \cite{shi-etal-2024-direct} adapts direct preference optimization to multi-turn trajectory optimization, 
and WKM \cite{qiao2024agent} regularizes actions with an external world knowledge model.
Similarly, KnowAgent \cite{zhu-etal-2025-knowagent} teaches LLM agents for self action learning from a knowledge base \cite{xia-etal-2025-knowledge} and NAT \cite{wang-etal-2025-nat} incorporates failure trajectories for finetuning with an adapted prompt prefix.
More recently, MPO \cite{xiong2025mpo} trains a meta planner agent that guides task execution agents.
Several agent tuning benchmarks and datasets have also emerged \cite{chen2024agent, song2024agentbank}.
In contrast, our proposed SA\textsc{n}D framework aims to teach LLM agents to effectively deliberate over candidate actions for better decision making.












\subsection{Deliberative Reasoning}

Prompting strategies for LLM deliberative reasoning have evolved rapidly. 
Chain-of-thought (CoT) prompting \cite{wei2022chain, xia-etal-2025-beyond} first showed that eliciting explicit intermediate reasoning steps markedly improves mathematical and symbolic reasoning. 
Building on this idea, ReAct blends CoT with environment feedback to couple reasoning and acting \cite{yao2023react}, while Self-Refine \cite{madaan2023self} and Reflexion \cite{shinn2023reflexion} introduce iterative self-critique loops that rewrite faulty thoughts. 
Tree-of-Thought \cite{yao2023tree} generalizes CoT into a breadth-first search over alternative thought branches, allowing the model to back-track and globally evaluate solutions.
SWAP \cite{xiong2024deliberate} frames deliberate reasoning as structure-aware planning with an internal world model.
\citet{guan2024deliberative} propose an explicit deliberation controller that decides when to generate, inspect or discard thoughts and \citet{karanam2024towards} study how many forward simulations are needed for reliable look-ahead in RL-style agents. 
Our SAnD framework extends the deliberative reasoning to LLM agent tasks with a focus of action deliberation.





\begin{figure*}[!t]
    \centering
    {\includegraphics[width=0.95\textwidth]{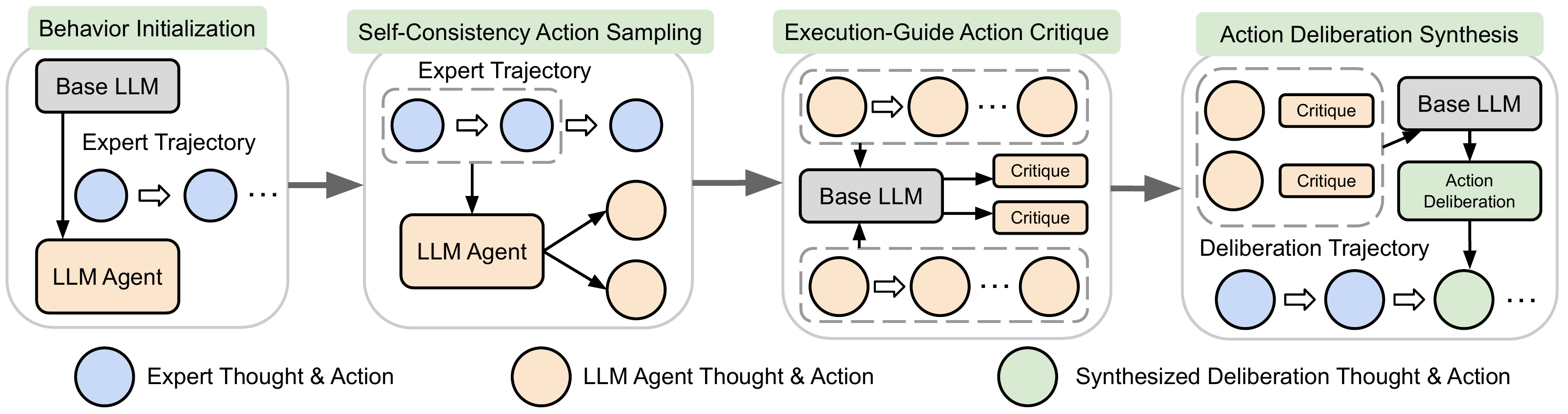}}
    \caption{An illustration of our SA\textsc{n}D framework for synthesizing one step of action deliberation thoughts.}
    \label{fig:overview} 
    \vspace{-1em}
\end{figure*}

\subsection{Iterative Self Learning}

Another relevant line of works enable a model to improve by repeatedly generating data and finetuning on its own synthesized output \cite{xia-etal-2025-selection}. 
The idea began with STaR \cite{zelikman2022star}, which bootstraps a few verified solutions into a large corpus of correct rationales.
RFT \cite{yuan2023scaling} generalises this to rejection-sampling proofs that pass an external checker. 
Subsequent work replaces hard filtering with self-feedback, e.g., Self-Refine \cite{madaan2023self} and SELF \cite{chen2023fireact} alternate draft–critique–revise loops.
Agent-R \cite{yuan2025agent} repairs failed trajectories via Monte-Carlo search before re-training, 
and \citet{karanam2024towards} show that only a handful of such self-play iterations are needed before returns saturate. 
Our SA\textsc{n}D framework follows the similar iterative self-learning idea to steadily improve LLM agents without additional human supervision.







%% file: 3-methodology.tex
\section{Task Formulation}

We formulate our studied agent tasks as multi-turn interactions between an LLM agent and a text-based environment following \citet{song-etal-2024-trial} and \citet{xiong2025mpo}.  
Specifically, for a ReAct-style \cite{yao2023react} LLM agent, the task begins with an instruction $u\in\mathcal{U}$.  
At each step, the LLM agent generates a reasoning thought $z\in\mathcal{Z}$ and an action $a\in\mathcal{A}$.  
The environment then returns an observation $o\in\mathcal{O}$.
At time step $t$, for an LLM agent $\pi_{\theta}$ with the past interaction history up to time step $t-1$ denoted as $h_{t-1} = (u,z_1,a_1,o_1,\dots,o_{t-1})$, the reasoning thought is sampled conditioned on the interaction history $z_t\sim\pi_{\theta}(\cdot\mid h_{t-1})$ followed by the action $a_t \sim \pi_{\theta}(\cdot\mid h_{t-1}, z_t)$.
Therefore, for a complete agent trajectory with $L$ steps $e = (u,z_1,a_1,o_1,\dots,o_{L-1}, z_L, a_L)$, the probability of generating it is given by
\begin{equation}
    \pi_{\theta}(e\mid u)\;=\;\prod_{t=1}^{L}
        \pi_{\theta}\bigl(z_t,a_t \mid h_{t-1}\bigr).
\end{equation}
After the task episode terminates upon success or maximum steps, the environment returns a task score
$r(u,e)\in[0,1]$ as the task successful rate.

\section{Methodology}
In this section, we describe in details our proposed Self-taught Actio\textsc{n} Deliberation (SA\textsc{n}D) framework.
Starting from a base LLM, SA\textsc{n}D iteratively finetunes it to be a stronger LLM agent using the deliberation thoughts generated by the base version of itself.
An intuitive illustration of our framework for generating a single step of deliberation thoughts can be found in Figure \ref{fig:overview}.
A more comprehensive overview of the entire iterative self-learning pipeline are presented in Algorithm \ref{algo:sand}.

\begin{algorithm*}[t!]
\DontPrintSemicolon
\KwIn{$\mathcal{D}_{\text{exp}}=\{(u,z_1,a_1,o_1,\dots,o_{L-1},z_L,a_L)^{(i)}\}$: expert trajectories, 
      $I$: number of self-taught iterations,
      $N$: number of sampled actions,
      $\pi_{\text{base}}$: base LLM,
      $\pi_{\theta} = \pi_{\text{base}} $: trainable LLM.}
\KwOut{Final LLM agent $\pi_{\theta}$}

Finetune $\pi_{\theta}$ on $\mathcal{D}_{\text{exp}}$:
$\displaystyle \mathcal{L}_{\mathrm{SFT}}=-\mathbb{E}_{e\sim\mathcal{D}_{\text{exp}}}\bigl[\log \pi_{\theta}(e\mid u)\bigr]$

\For{$k = 1$ \KwTo $I$}{
    $\pi_{k}\leftarrow\pi_{\theta}$, $\mathcal{D}_{\text{delib}}\leftarrow\emptyset$\;
\ForEach{$e=(u,z_1,a_1,o_1,\dots,z_L,a_L)\in\mathcal{D}_{\text{exp}}$}{
  Initialize history $h_0\leftarrow u$ and self-taught deliberation trajectory $\tilde{e} = (u)$\;
  \For{$t=1$ \KwTo $L$}{
     Sample $N$ actions:
      $\{\hat{z}_t^{(n)},\hat{a}_t^{(n)}\}_{n=1}^{N}\sim\pi_k(\cdot\mid h_{t-1})$\;
      \lIf{$\lvert\{\hat{a}_t^{(1)},\dots,\hat{a}_t^{(N)},a_t\}\rvert=0$}{\textbf{continue}}
    Rollout each action: $\{\hat{e}_t, r_t\} \sim \pi_{k}(\cdot \mid h_{t-1}, \hat{z}_t,\hat{a}_t)$\; Generate critique for each action: $    c_t
\sim
\pi_{\text{base}}(\cdot \,\bigm|\,
\hat{a}_t,\hat{e}_t,r_t, \texttt{Prompt}_c
),$\;
    Synthesize action deliberation thought: $\tilde{z}_t \sim \pi_{\text{base}}(\cdot \mid \{(\hat{a}_t^{(n)},\,c_t^{(n)})\}_{n=1}^{N+1}, \texttt{Prompt}_d)$ \;
    $\tilde{e}
  \leftarrow
  \tilde{e}\cup(\tilde{z}_t,a_t,o_t)$;
  $h_t
  \leftarrow
  (h_{t-1},z_t,a_t,o_t)$\;

  }
  $\mathcal{D}_{\text{delib}}\leftarrow\mathcal{D}_{\text{delib}}\cup\{\tilde{e}\}$\;
}

        Finetune $\pi_{\theta}$ on $\mathcal{D}_{\text{delib}}$:
        $\displaystyle \mathcal{L}_{\mathrm{SFT}}
        = -\mathbb{E}_{\tilde{e}\sim\mathcal{D}_{\text{delib}}}
        \bigl[\log \pi_{\theta}(\tilde{e}\mid u)\bigr]$
    
    Set $\mathcal{D}_{\text{exp}}\leftarrow\mathcal{D}_{\text{delib}}$ for the next iteration\;
}
\Return $\pi_{\theta}$
\caption{\textbf{Self-Taught Action Deliberation (SA\textsc{n}D)}}
\label{algo:sand}
\end{algorithm*}

\subsection{Behavior Initialization}
We start from a base instruction-tuned LLM $\pi_{\text{base}}$.  
Following \citet{song-etal-2024-trial} and \citet{xiong-etal-2024-watch}, we initialize an LLM agent with the basic reasoning and action behavior for completing the task via supervised finetuning (SFT) on a set of ReAct-style expert trajectories on training tasks
$\mathcal{D}_{\text{exp}}=\{(u,e)^{(i)}\}_{i=1}^{|\mathcal{D}|}$ with the loss
\begin{equation}
\mathcal{L}_{\text{SFT}}
=
-
\mathbb{E}_{e\sim\mathcal{D}_{\text{exp}}}
\bigl[\log \pi_{\theta}(e\mid u)\bigr].
\end{equation}
We then obtain the initial LLM agent policy $\pi_{\theta}$ for the subsequent iterative improvement.  

\subsection{Self-Consistency Action Sampling}\label{sec:sampling}
With an LLM agent policy $\pi_{\theta}$, we aim to further teach agent the action deliberation behavior.
Two central questions here are (i) when the agent should invest extra thinking over actions and (ii) what actions to think about, especially within a large or even unbounded action space.
To address them, we utilize self-consistency action sampling which offers a natural solution. 

For each expert trajectory $e$, we replay every expert interaction and branch at each step $t$.  
Specifically, given expert interaction history $h_{t-1}$, the current policy $\pi_{\theta}$ samples $N$ actions
\begin{equation}
    \{\hat{a}_t^{(1)},\dots,\hat{a}_t^{(N)}\}\sim\pi_{\theta}(\cdot\mid h_{t-1}),
\end{equation}
where we omit the sampled reasoning thoughts $\hat{z}_t$ here for notation simplicity.
Together with the original expert action $a_t$, we form a candidate action set of size $N+1$.  

We then define an inconsistency indicator that flags whether deliberation is needed for step $t$:
\begin{equation}
    \mathbf{1}_{\text{delib}}(t)
=
\mathbf{1}\Bigl(\bigl|\{\hat{a}_t^{(1)},\dots,\hat{a}_t^{(N)},a_t\}\bigr|>1\Bigr).
\end{equation}\label{eq:flag}
If all actions in the set are the same, $\mathbf{1}_{\text{delib}}(t)=0$, showing that the predictive distribution $\pi_{\theta}(\cdot\mid h_{t-1})$ is sharply peaked, this suggests that the model is confident in conducting the expert action $a_t$ or the decision at the current state is trivial.
In this case, no extra reasoning or deliberation is needed.  
When the set contains more than one unique action, $\mathbf{1}_{\text{delib}}(t)=1$, this suggests the uncertainty of the LLM agent at the current state, and generating an explicit deliberation thought can help the agent better choose among candidate actions.

Moreover, since every branch starts from a step on the expert trajectory $e$, the sampled actions $\hat{a}_t$ remain close to both the demonstration distribution and the current LLM policy distribution while still exploring diverse futures, thereby avoiding random exploration over the large action space.


\subsection{Execution-Guided Action Critique}\label{sec:eac}
If the inconsistency indicator flags for action deliberation at step $t$, $\mathbf{1}_{\text{delib}}(t)=1$, then next question is how LLM agent can learn to generate meaningful step-level action evaluations when deliberating over the candidate set.
In typical multi-turn interaction tasks, the reward is often delayed till task completion \cite{xia-etal-2024-aligning, zhang2025survey}. 
Therefore, to provide additional context and evaluation signals for each candidate action, we collect its full rollout by executing each action $\hat{e}_t \sim \pi_{\theta}(\cdot \mid h_{t-1},\hat{a}_t)$ and obtain the final task reward $r_t\!\in\![0,1]$ from the training environment.

Then, for each candidate action rollout, we prompt the frozen base LLM to generate a verbal critique $c_t$ of the candidate action $\hat{a}_t$ guided by its execution results $\hat{e}_t$ and $r_t$
\begin{equation}
    c_t
\sim
\pi_{\text{base}}\bigl(\cdot \,\bigm|\,
\hat{a}_t,\hat{e}_t,r_t, \texttt{Prompt}_c
\bigl),
\end{equation}
where $\texttt{Prompt}_c$ is the critique prompt detailed in Figure \ref{fig:action_critique_prompt}. 
It shows the action, the ensuing sequence of observations, and the final reward, and asks for a concise verdict that states whether the action advanced, hindered, or had no effect on task success.
As the critique is verbalized natural language, we also specify in the prompt for the base LLM to record reusable commonsense knowledge (e.g., ``eggs are more likely to be stored in the refrigerators'') that is not tied to the specific task instance.
Such commonsense snippets accumulate across rollouts and provide transferable cues for more informative step-level action evaluation than numerical values aggregated from Monte Carlo rollouts \cite{xiong-etal-2024-watch,lin2025qlass}.

\subsection{Action Deliberation Synthesis}\label{sec:delibsyn}
After all critiques $c_t^{(n)}$ on candidate actions $\hat{a}_t^{(n)}$ have been gathered, we prompt the base LLM $\pi_{\text{base}}$ to generate a single deliberation thought. 
The prompt, detailed in Figure \ref{fig:action_deliberation_prompt}, instructs the LLM to first propose and analyze each candidate action explicitly, then compare over them, and give a rationale for the final action choice of the expert action $a_t$ at the current step 
\begin{equation}
    \tilde{z}_t
\sim
\pi_{\text{base}}
\bigl(\cdot \bigm|\,
\{(\hat{a}_t^{(n)},c_t^{(n)})\}_{n=1}^{N+1}, \texttt{Prompt}_d
\bigr).
\end{equation}
We then append $(\tilde{z}_t,a_t,o_t)$ to the self-augmented deliberation trajectory $\tilde{e}$ collected along each step and update the running history $h_t$.

Note that we keep the expert action $a_t$ as the ground-truth action here assuming it is the optimal one at the current step.
However, as some expert data is annotated by human or another LLM, the LLM agent being finetuned may explore better paths than the expert path \cite{song-etal-2024-trial, xiong-etal-2024-watch}. 
Thus, we also devise an optional expert switch mechanism that replaces the original expert action with a better explored action if the LLM agent finds a better rollout during execution in Section \ref{sec:eac}.

\subsection{Iterative Deliberation Finetuning}
Exploring through all training tasks, the collection of self-taught action deliberation trajectories is denoted by $\mathcal{D}_{\text{delib}}=\{(u,\tilde{e})^{(i)}\}_{i=1}^{|\mathcal{D}_{\text{delib}}|}$.
We update the LLM agent $\pi_{\theta}$ with via the similar supervised finetuning objective
\begin{equation}
    \mathcal{L}_{\text{SFT}}
=
-
\mathbb{E}_{\tilde{e}\sim\mathcal{D}_{\text{delib}}}
\bigl[\log \pi_{\theta}(\tilde{e}\mid u)\bigr].
\end{equation}
Compared with the initial expert trajectories, the synthesized deliberation trajectories provide richer guidance on enabling the action deliberation behavior as well as on why an action is chosen among alternative candidates, rather than only what action to mimic.
Moreover, as the action deliberation is synthesized only when the action inconsistency indicator $t$, $\mathbf{1}_{\text{delib}}(t)=1$ defined in Equation \ref{eq:flag} flags, the trajectories $\mathcal{D}_{\text{delib}}$ we collected are mixed with deliberation and non-deliberation steps.
This also teaches the LLM agent when to conduct action deliberation, as justified by our empirical analysis discussed in Section \ref{sec:delib_rate}.
Note that the LLM agent finetuned on the deliberation trajectories does not perform any action sampling during inference time. Instead, it generates the entire action deliberation thought in one pass, as illustrated in Figure \ref{fig:intro}.

Finally, we set $\mathcal{D}_{\text{exp}}\!\leftarrow\!\mathcal{D}_{\text{delib}}$ and repeat the sampling, critique, synthesis, and finetuning loop for $I$ iterations, steadily improving LLM agents with a base version of itself without additional human labels or annotations.

%% file: 4-experiment.tex
\section{Experimental Setup}\label{sec:setup}

\subsection{Datasets and Evaluation}
We evaluate our proposed SA\textsc{n}D agent tuning framework mainly in two representative interactive environments \textbf{ALFWorld} and \textbf{ScienceWorld} following \citet{xiong2025mpo}. 
ALFWorld \cite{shridhar2020alfworld} provides a text-based household task environment that for natural language understanding and embodied reasoning. 
It provides only binary rewards of task success upon completion or termination. 
ScienceWorld \cite{wang2022scienceworld} presents a text-based environment where agents perform elementary-level scientific experiments. 
It offers a granular reward system that quantifies partial progress toward scientific task goals.
Both datasets include training sets and test sets for both seen and unseen tasks as reported in Table \ref{tab:data}, allowing us to assess how well LLM agents finetuned with SA\textsc{n}D can generalize to unseen scenarios.
We also report additional evaluation results on a real-world web navigation task WebShop \cite{yao2022webshop} in Appendix \ref{app:webshop}.
Following \citet{song-etal-2024-trial} and \citet{xiong-etal-2024-watch}, we use the \textbf{Average Reward} across test tasks as our main evaluation metric. 
We set the decoding temperature to 0 for all agents when evaluating on the test sets to facilitate reproducibility.

\begin{table}[t]
    \centering
    \resizebox{\linewidth}{!}{
    \begin{tabular}{l c c c c}
    \toprule
    \textbf{Dataset}   & \textbf{Train} & \textbf{Test Seen} & \textbf{Test Unseen} & \textbf{Action Space}\\
    \midrule
    ScienceWorld & 1483 & 194 & 211 & 19\\
    ALFWorld & 3321 & 140 & 134 & 13\\
    \bottomrule
    \end{tabular}
    }
    \caption{Statistics of ALFWorld and SciWorld datasets.}
    \label{tab:data}
    \vspace{-1em}
\end{table}

\begin{table*}[!t]
    \centering
    \resizebox{\linewidth}{!}{
    \begin{tabular}{l c | c c c c | c}
    \toprule
    \multirow{2}{*}{\textbf{Model}} & \multirow{2}{*}{\textbf{Single Agent}}  & \multicolumn{2}{c}{\textbf{ScienceWorld}} & \multicolumn{2}{c|}{\textbf{ALFWorld}} & \multirow{2}{*}{\textbf{Average}}  \\ \cmidrule(l){3-4} \cmidrule(l){5-6}
    & & Seen & Unseen & Seen & Unseen & \\ \midrule
    \multicolumn{7}{l}{\textit{Agents w/o Training}} \\ \midrule
    GPT-4o~\citep{achiam2023gpt} & \ding{51} & 60.0 & 56.0 & 78.6 & 83.6 & 69.6 \\
    GPT-4o-mini~\citep{achiam2023gpt} & \ding{51} & 49.1 & 42.7 & 32.1 & 41.0 & 41.2 \\
    Llama-3.1-8B-Instruct~\citep{dubey2024llama} & \ding{51} & 47.7 & 42.2 & 22.9 & 28.4 & 35.3 \\
    Llama-3.1-8B-Instruct + MPO \cite{xiong2025mpo} & \ding{55} & 56.5 & 55.5 & 50.0 & 52.2 & 53.6
\\
    Qwen2.5-7B-Instruct~\citep{yang2025qwen3} & \ding{51} & 38.5 & 38.8 & 71.4 & 75.4 & 56.0 \\
    Llama-3.1-70B-Instruct~\citep{dubey2024llama} & \ding{51} & 72.6 & 70.2 & 78.6 & 73.9 & 73.8 \\ 
    Llama-3.1-70B-Instruct + MPO \cite{xiong2025mpo} & \ding{55} & {80.4}  & \underline{79.5} & 85.7 & 86.6 & {83.1} \\ \midrule
    \multicolumn{7}{l}{\textit{Agents w/ Training}} \\ \midrule
    Qwen2.5-7B-Instruct + SFT~\citep{zeng-etal-2024-agenttuning} & \ding{51} & 69.2 & 60.8 & 72.1 & 75.4 & 69.4 \\
    Llama-3.1-8B-Instruct + SFT~\citep{zeng-etal-2024-agenttuning} & \ding{51} & 75.6 & 65.1 & 79.3 & 71.6 & 72.9 \\
    Llama-3.1-8B-Instruct + ETO~\citep{song-etal-2024-trial} & \ding{51} & 81.3 & 74.1 & 77.1 & 76.4 & 77.2 \\
    Llama-3.1-8B-Instruct + KnowAgent~\citep{zhu-etal-2025-knowagent} & \ding{51} & 81.7 & 69.6 & 80.0 & 74.9 & 76.6 \\
    Llama-3.1-8B-Instruct + WKM~\citep{qiao2024agent} & \ding{55} & 82.1 & 76.5 & 77.1 & 78.2 & 78.5 \\
    Llama-3.1-8B-Instruct + ETO\&MPO \cite{xiong2025mpo} & \ding{55} & {83.4} & \textbf{80.8} & {85.0} & 79.1 & {82.1} \\ \midrule
    \rowcolor[RGB]{242, 255, 253}Qwen2.5-7B-Instruct + SA\textsc{n}D (Iteration 1) & \ding{51} & 80.9 & 67.2 & 85.7 & 85.0 & 79.7 \\
    \rowcolor[RGB]{242, 255, 253}Qwen2.5-7B-Instruct + SA\textsc{n}D (Iteration 2) & \ding{51} & 83.2 & 69.9 & 85.0 & 89.6 & 81.9 \\
    \rowcolor[RGB]{242, 255, 253}Qwen2.5-7B-Instruct + SA\textsc{n}D (Iteration 3) & \ding{51} & 84.0 & 69.0 & 90.7 & 94.8 & 84.6 \\
    \rowcolor[RGB]{242, 255, 253}Llama-3.1-8B-Instruct + SA\textsc{n}D (Iteration 1) & \ding{51} & \underline{86.6} & 77.5 & \underline{92.9} & 91.8 & 86.0 \\
    \rowcolor[RGB]{242, 255, 253}Llama-3.1-8B-Instruct + SA\textsc{n}D (Iteration 2) & \ding{51} & \textbf{88.7} & 78.2 & \textbf{94.3} & \underline{94.0} & \underline{88.8} \\
    \rowcolor[RGB]{242, 255, 253}Llama-3.1-8B-Instruct + SA\textsc{n}D (Iteration 3) & \ding{51} & {85.7} & {79.1} & \textbf{94.3} & \textbf{96.3} & \textbf{88.9} \\
    \bottomrule
    \end{tabular}
    }
    \caption{Average rewards of all compared methods on two datasets. SA\textsc{n}D significantly improves LLM agents across different model backbones, outperforming proprietary LLMs as well as state-of-the-art multi-agent approaches.}
    \label{tab:main_results}
\end{table*}

\subsection{Baselines and Variants}
We compare SA\textsc{n}D with the following agent tuning baselines and variants
\begin{itemize}[left=0pt, parsep=0pt, partopsep=0pt]
    \item \textbf{AgentTuning} \cite{zeng-etal-2024-agenttuning}: a direct supervised finetuning approach on expert trajectories.
    \item \textbf{ETO} \cite{song-etal-2024-trial}: a representative agent tuning method leveraging an adapted direct preference optimization objective for contrastive agent trajectory pairs.
    \item \textbf{KnowAgent} \cite{zhu-etal-2025-knowagent}: a recent framework employing an additional action knowledge base for self learning of LLM agents.
    \item \textbf{WKM} \cite{qiao2024agent}: an agent tuning method with a jointly optimized world knowledge model available during test time.
    \item \textbf{MPO} \cite{xiong2025mpo}: an optimization approach via training a meta planner agent generating explicit guidance for task execution agents.
    \item \textbf{SA\textsc{n}D}$_{\text{w/o\;SAS}}$: a variant of our method which does not conduct self-consistency action sampling (SAS) but instead directly prompts the base LLM to generate $N$ alternative candidate actions in context during action deliberation synthesis.
    \item \textbf{SA\textsc{n}D}$_{\text{w/o\;EAC}}$: a variant of our method which skips the execution-guided action critique (EAC) stage and directly synthesize action deliberation thought with $N$ sampled candidate actions.
\end{itemize}
For more comprehensive comparison, we also report results of prompting-based ReAct-style LLM agent based on proprietary and open-sourced models \textbf{GPT4o} \cite{achiam2023gpt} and \textbf{Llama-3.1-70B-Instruct} \citep{dubey2024llama} collected by \citet{xiong2025mpo}, where an in-context example is given for all prompting-based models. 
We provide in Appendix \ref{app:reward} additional discussion and comparisons of our SA\textsc{n}D framework with recent test-time search methods guided by process reward or Q-value models \cite{zhai2025enhancing, lin2025qlass, xia2025agentrm}.


\begin{figure*}[t!]
    \centering
    {\includegraphics[width=0.251\textwidth]{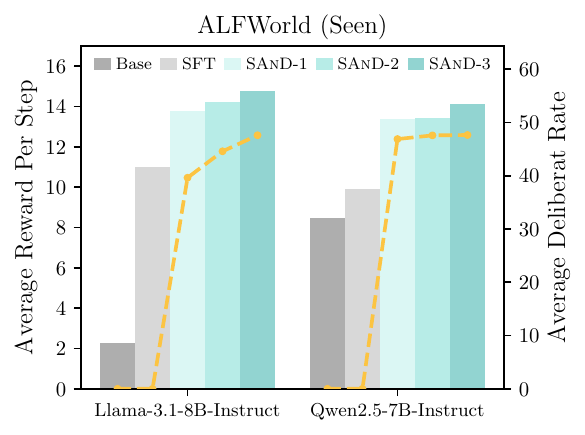}\hspace{-0.3em}}
    {\includegraphics[width=0.251\textwidth]{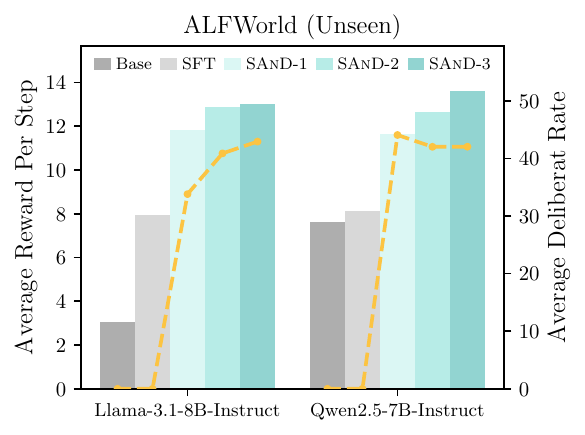}\hspace{-0.3em}}
    {\includegraphics[width=0.251\textwidth]{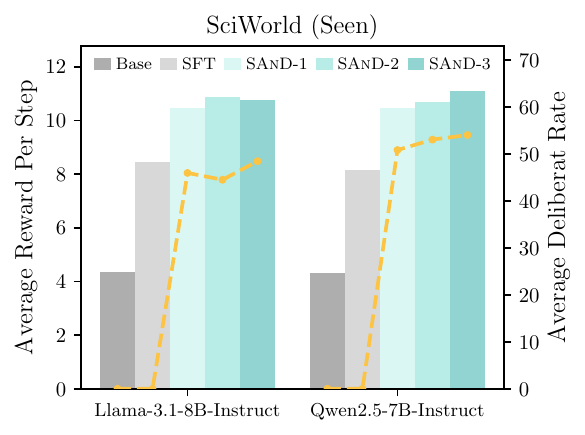}\hspace{-0.3em}}
    {\includegraphics[width=0.251\textwidth]{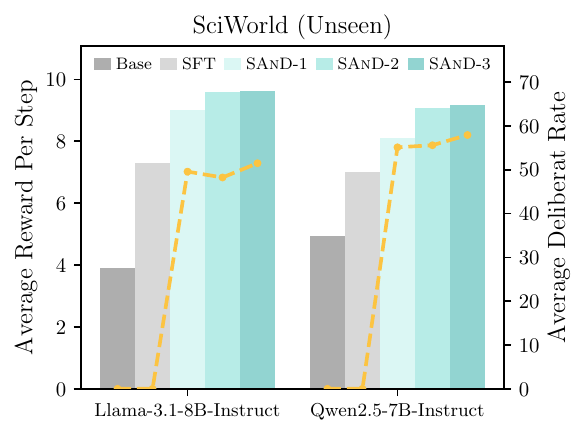}\hspace{-0.3em}}
    \vspace{-0.5em}
    \caption{Average reward per step (bars) and average action deliberation rate per step (lines) on test sets.}
    \label{fig:avg_reward_step} 
    \vspace{-0.5em}
\end{figure*}

\subsection{Implementation Details}\label{sec:details}
We adopt two backbone models Llama-3.1-8B-Instruct \cite{dubey2024llama} and Qwen2.5-7B-Instruct \cite{yang2025qwen3} as the base models and finetune them with our SA\textsc{n}D framework.
The initial expert trajectories are collected by \citet{song-etal-2024-trial}.  
For behavior initialization step, we follow \citet{song-etal-2024-trial} to set batch size of 64 with a learning rate of 1e-5 and a cosine scheduler for 3 epochs.
At self-consistency action sampling step, the decoding temperature of the LLM agent $\pi_\theta$ is set to 1.0 for sampling $N = 5$ candidate actions as well as the subsequent rollout execution.
The execution-guided action critique is generated by the base LLM $\pi_{\text{base}}$ with the decoding temperature 0. 
Both prompts for critique generation and action deliberation synthesis are provided in Appendix \ref{app:prompt}.
We disable the expert action switch mechanism discussed in Section \ref{sec:delibsyn} on ScienceWorld as we empirically observe that some of the tasks have short-cuts that might boost LLM agents on training set but hurt performances on test set.
For deliberation finetuning steps, we set similarly batch size of 64 and learning rate of 1e-5 for $I=3$ iterations. 
To avoid overfitting, we train 3 epochs only for the first iteration of SA\textsc{n}D and 1 epoch for later iterations.
We use OpenRLHF \cite{hu2024openrlhf} to implement our training framework and all experiments run on 8 NVIDIA A100 80GB GPUs.


%% file: 5-result.tex
\section{Results}

\begin{figure*}[!t]
    \centering
    {\includegraphics[width=1\textwidth]{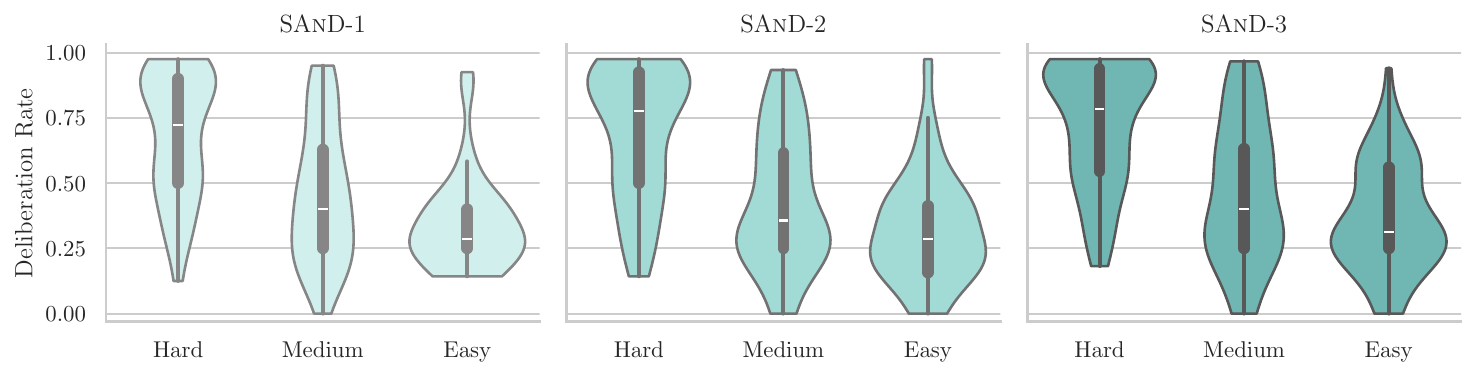}\label{fig:n_amazon}\hspace{-0.3em}}

    \vspace{-0.5em}
    \caption{Action deliberation rate distribution across three difficulty bands in unseen test set on ScienceWorld.
    Each panel corresponds to a SA\textsc{n}D iteration starting from Llama-3.1-8B-Instruct. 
    The difficulty bands \textit{Hard}, \textit{Medium}, \textit{Easy} are determined based on the tertiles of reward distribution from the base Llama-3.1-8B-Instruct.
    The results show that more SA\textsc{n}D iterations teach LLM agents to deliberate more on hard tasks and less on easy tasks.}
    \label{fig:delib_rate} 
    \vspace{-0.5em}
\end{figure*}

\subsection{How does SA\textsc{n}D perform compared with other agent tuning methods?}

We show the results of all compared methods on both seen and unseen test tasks in Table \ref{tab:main_results}. 
From the results, we observe a clear advantage of SA\textsc{n}D which outperforms all baselines on ALFWorld by a large margin.
On ScienceWorld, SA\textsc{n}D also shows competitive performances matching or surpassing state-of-the-art multi-agent approach.
For both Llama-3.1-8B-Instruct and Qwen-2.5-7B-Instruct as the backbone LLMs, SA\textsc{n}D (Iteration 3) achieves an average over 20\% performance boost compared with SFT on initial expert data.  

Besides, with our iterative deliberation finetuning, we also observe a steady performance improvement across different iterations of SA\textsc{n}D, demonstrating the effectiveness of our self learning framework requiring no additional human labels.
Another notable observation is that on later iterations of SA\textsc{n}D, agents trained on both Llama-3.1-8B-Instruct and Qwen-2.5-7B-Instruct exhibit strong generalization capabilities on ALFWorld unseen tasks, achieving high rewards even than seen tasks.
We attribute the performance gains on unseen tasks to the action deliberation behavior learned by LLM agents during SA\textsc{n}D iterations. 
Such action deliberation behavior enables LLM agents to explicitly analyze unseen actions and environments before committing one instead of relying mostly on seen action patterns learned during training tasks.

\begin{table}[!t]
    \centering
    \resizebox{0.8\linewidth}{!}{
    \begin{tabular}{l | c c | c c}
    \toprule
    \multirow{2}{*}{\textbf{Method}}  & \multicolumn{2}{c}{\textbf{ScienceWorld}} & \multicolumn{2}{c}{\textbf{ALFWorld}}\\ 
    & Seen & Unseen & Seen & Unseen \\
    \midrule
    \multicolumn{5}{l}{\textit{Qwen2.5-7B-Instruct}}\\
    \midrule
    Base & 38.5 & 38.8 & 71.4 & \underline{75.4} \\
    SFT & 69.2 & 60.8 & \underline{72.1} & \underline{75.4} \\
    SA\textsc{n}D$_{\text{w/o\;SAS}}$ & 63.5 & 52.4 & \underline{72.1} & 62.7 \\
    SA\textsc{n}D$_{\text{w/o\;EAC}}$ & \underline{72.0} & \underline{66.3} & 70.6 & 75.0 \\
    SA\textsc{n}D & \textbf{80.9} & \textbf{67.2} & \textbf{85.7} & \textbf{85.0} \\ \midrule
    \multicolumn{5}{l}{\textit{Llama-3.1-8B-Instruct}}\\
    \midrule
    Base & 47.7 & 42.2 & 22.9 & 28.4 \\
    SFT & 75.6 & 65.1 & 79.3 & 71.6 \\
    SA\textsc{n}D$_{\text{w/o\;SAS}}$ & 70.3 & 62.0 & \underline{85.7} & 77.3 \\
    SA\textsc{n}D$_{\text{w/o\;EAC}}$ & \underline{78.6} & \underline{73.7} & 85.0 & \underline{86.6} \\
    SA\textsc{n}D & \textbf{86.6} & \textbf{77.5} & \textbf{92.9} & \textbf{91.8} \\ 
    \bottomrule
    \end{tabular}
    }
    \caption{Ablation study on different modules in SA\textsc{n}D.}
    \label{tab:ablation}
    \vspace{-1em}
\end{table}

\subsection{Are self-consistency action sampling and execution-guided critique necessary?}

To validate the effectiveness of our devised self-consistency action sampling and execution-guided action critique, we compare {SA\textsc{n}D} at the first iteration with its ablated variants {SA\textsc{n}D}$_{\text{w/o\;SAS}}$ and {SA\textsc{n}D}$_{\text{w/o\;EAC}}$.
The results are shown in Table \ref{tab:ablation}, where we observe a performance drop after removing each modules.
Specifically, we find that {SA\textsc{n}D}$_{\text{w/o\;SAS}}$ can even hurt the agent performance being outperformed by initial SFT. 
From our logged failed testing trajectories, we observe that without self-consistency action sampling, LLM agents often propose random actions irrelevant to the task goals and sometimes show degenerated behavior of repeating a candidate action till the maximum context length.
On the other hand, {SA\textsc{n}D}$_{\text{w/o\;EAC}}$, though also showing a small performance decrease compared with {SA\textsc{n}D}, still improves over the initial SFT agent.
The results again demonstrates the necessity of the self-consistency action sampling module while also validating the effectiveness of execution-guided action critique in improving the synthesized deliberation quality.

\subsection{Does action deliberation improve LLM agents at step-level across iterations?}
SA\textsc{n}D has shown overall performance improvement over iterations in Table \ref{tab:main_results}. To further study the influence of the action deliberation behavior LLM agents learned from SA\textsc{n}D, we show in Figure \ref{fig:avg_reward_step} the average reward per step and the corresponding average action deliberation rate per step across all test sets.
The per-step average reward is calculated as the ratio of final reward to the total steps for each task, averaged across all tasks in the test set.
Similarly, the per-step average deliberation rate is the ratio of action deliberation steps to the total steps for each task, averaged across all tasks.

From Figure \ref{fig:avg_reward_step}, we can first observe a consistent improvement on per-step average reward across different finetuning iterations with first iteration shows a larger gain followed by smaller gains in later iterations.
We also observe that the per-step action deliberation rate also show a general increasing pattern.
Such correlation further validates the advantage of step-level action deliberation, which enables LLM agent to make better decisions at each step.
The higher step-level reward also brings the advantages of earlier and more efficient task completion for practical applications of LLM agents.

\subsection{Do LLM agents finetuned with SA\textsc{n}D really learn when to deliberate?}\label{sec:delib_rate}

To further analyze the agent tuning dynamics during SA\textsc{n}D iterations, we study whether LLM agents have learned to decide when to deliberate over candidate actions, as discussed in Section \ref{sec:sampling}.
Specifically, we visualize when the LLM agent decides to deliberate with violin plots in Figure \ref{fig:delib_rate}, where each panel corresponds to an iteration in SA\textsc{n}D.
As ScienceWorld provides finegrained rewards that can reflect partial task completion rate, we partition the unseen tasks on ScienceWorld into three difficulty bands based on the empirical tertiles of reward distribution from the base LLM Llama-3.1-8B-Instruct.
We define the bottom third as \textit{Hard} tasks, the middle third as \textit{Medium} tasks, and the top third as \textit{Easy} tasks. 
Within each band we compute the deliberation rate of SA\textsc{n}D similarly defined as the ratio of deliberation steps to the total steps for each task, and plot the distribution of deliberation rates across tasks.

From Figure \ref{fig:delib_rate}, we observe that across all three iterations the hard band remains the only one with a high median deliberation rate around 0.75, while the median deliberation rate on easy band stays near 0.30.
This shows SA\textsc{n}D effectively teaches LLM agent to direct more action deliberation to hard tasks while keeping reasoning concise when the task is easy.
From iteration 1 to iteration 3, we also observe a slight distribution shift of the hard violin, which widens at the top with the median gradually increases.
This further demonstrates the effectiveness of iterative deliberation finetuning in our SA\textsc{n}D framework that not only improves the task performances but also teaches LLM agents to make better decisions on when to deliberate.

\begin{table}[!t]
    \centering
    \resizebox{0.85\linewidth}{!}{
    \begin{tabular}{l | c c}
    \toprule
    \textbf{Method} & \textbf{ALFWorld} & \textbf{ScienceWorld} \\
    \midrule
    SFT & 498.3 & 800.0 \\
    SAND (Iteration 1) & 1,314.2 (2.6$\times$) & 2,411.9 (3.0$\times$) \\
    SAND (Iteration 2) & 1,105.8 (2.2$\times$) & 2,522.1 (3.2$\times$) \\
    SAND (Iteration 3) & 1,146.2 (2.3$\times$) & 2,253.6 (2.8$\times$) \\
    \bottomrule
    \end{tabular}
    }
    \caption{Average \#tokens per task on ALFWorld and ScienceWorld. Multipliers are relative to SFT agent.}
    \label{tab:cost}
    \vspace{-1em}
\end{table}

\subsection{How much additional inference-time computation cost does SA\textsc{n}D introduce?}\label{sec:cost}
As SA\textsc{n}D teaches LLM agents to explicitly deliberate over candidate actions, it introduces additional computation cost during inference time.
To study how much additional inference-time cost is incurred, we compare in Table~\ref{tab:cost} the average number of tokens used per task between the SFT agent (without action deliberation) and our SA\textsc{n}D-finetuned agents (with action deliberation), where the base model is Llama-3.1-8B-Instruct.

From Table~\ref{tab:cost}, we find that the additional action deliberation results in approximately 2 to 3 times more tokens per task.
Compared to representative test-time scaling approaches such as Best-of-N, which incurs 5 times more tokens when $N=5$, we believe our SA\textsc{n}D framework introduces a reasonable additional inference-time computation cost with considerable performance improvements.
Moreover, as analyzed in Section~\ref{sec:delib_rate}, our SA\textsc{n}D framework effectively teaches LLM agents when to deliberate, avoiding unnecessary action deliberation on simple tasks.
This finding is also reflected in Table~\ref{tab:cost}, where a slight decreasing trend in token usage is observed across iterations, indicating better inference-time computation usage through our iterative finetuning framework.


%% file: 6-conclusion.tex
\section{Conclusion}

In this paper, we propose {Self-taught Actio\textsc{n} Deliberation} (SA\textsc{n}D), a self-learning framework that equips LLM agents with explicit action deliberation. 
Addressing when and what to deliberate given large action space, SA\textsc{n}D samples candidate actions by self-consistency, critiques each action guided exectured rollout, synthesizes a deliberation thought, and iteratively finetunes the LLM agent on the enriched trajectories.  
Experiments and analysis demonstrate the effectivenes and advantages of our methods, which further highlights the key role of deliberative reasoning in developing more powerful LLM agents for real world applications.

\section*{Limitations}
Despite the performance improvements, generating more deliberation thoughts inevitably increases the token usage and inference costs.  
As discussed and analyzed in Section \ref{sec:delib_rate}, our proposed SA\textsc{n}D framework teaches LLM agent when to deliberate via self-consistency action sampling to avoid deliberating during trivial decision making steps. 
Our results in Section \ref{sec:cost} further show that the action deliberation learned by SA\textsc{n}D introduces reasonable additional inference-time computation cost.
To further improve the reasoning efficiency, more advanced methods such reinforcement learning or direct preference optimization can be utilized to guide the LLM agent to better decide when to generating more comprehensive deliberative reasoning and when to generate more concise quick thoughts.
Parallel inference techniques can also be applied to further enhance the inference efficiency.

%% file: 7-appendix.tex
\section*{Appendix}

\begin{table*}[!t]
    \centering
    \renewcommand{\arraystretch}{2.25}
    \resizebox{\linewidth}{!}{
    \begin{tabular}{l | c c c | c c c}
    \toprule
    \textbf{Method} &
    \textbf{\makecell{Train Base \\ LLM Agent}} &
    \makecell{\textbf{Train Separate}\\\textbf{PRM/Value Model}} &
    \textbf{\makecell{Inference-time \\ Sampling Strategy}} &
    \textbf{WebShop} &
    \textbf{\makecell{ALFWorld \\(Unseen)}} &
    \textbf{\makecell{SciWorld \\(Unseen)}} \\
    \midrule
    \makecell[l]{Llama-3.1-8B-Instruct \\ \hspace{0.5cm} + Q~\cite{zhai2025enhancing}} & \ding{55} & \checkmark & 5 Actions Per Step & 60.0 & -- & -- \\
    \makecell[l]{Llama-2-7B-Chat \\ \hspace{0.5cm} + QLASS~\cite{lin2025qlass}}         & \checkmark & \checkmark & 6 Actions Per Step & 70.3 & 82.8 & 66.4 \\
    \makecell[l]{Llama-3-8B-Instruct \\ \hspace{0.5cm} + AgentRM-BoN~\cite{xia2025agentrm}}  & \checkmark & \checkmark & Best-of-5 Trajectories & 71.0 & 94.8 & 76.1 \\
    \makecell[l]{Llama-3-8B-Instruct \\ \hspace{0.5cm} + AgentRM-Beam~\cite{xia2025agentrm}} & \checkmark & \checkmark & \makecell{25 Actions Per Step \\ \texttt{(5$\times$5 Beam Search)}} & 75.3 & 96.3 & 82.6 \\
    \makecell[l]{Llama-3.1-8B-Instruct \\ \hspace{0.5cm} + SA\textsc{n}D (Ours)}        & \checkmark & \ding{55} & \makecell{1 Action Per Step \\ \texttt{(No Sampling)}} & 72.4 & 96.3 & 79.1 \\
    \bottomrule
    \end{tabular}
    }
    \caption{Comparisons of SA\textsc{n}D with representative test-time search methods guided by PRM or Q-value model.}
    \label{tab:prm}
\end{table*}

\section{Additional Results on Webshop}\label{app:webshop}
To further verify the generalizability of SAND to more diverse environments, we report in Table \ref{tab:webshop} the performance of SAND with Llama-3.1-8B-Instruct as the base model on a real-world web navigation task WebShop \cite{yao2022webshop}. 
We use the same train-test dataset splits as in \citet{song-etal-2024-trial}.
The number of sample actions in our self-consistency action sampling is set to $N=3$ due to the smaller action space of WebShop compared to ALFWorld and SciWorld.
Other configurations remain the same as in Section \ref{sec:details}.
From the results, we observe a consistent performance boost with our SA\textsc{n}D framework for LLM agents with around 10\% improvement compared to the SFT baseline, which validates the effectiveness of SA\textsc{n}D on more diverse environments.

\section{Comparisons with PRM and Q-Value Models for LLM Agents}\label{app:reward}

In this work, we propose an LLM agent tuning framework, SA\textsc{n}D, that enhances LLM agents' abilities during training time with self-taught deliberation trajectories.
During inference, our SA\textsc{n}D-finetuned LLM agent generates the entire action deliberation thought along with the final action in one pass, as illustrated in Figure~\ref{fig:intro}.
Therefore, our proposed LLM agent tuning framework is orthogonal and complementary to recent process reward model (PRM) or Q-value model-guided test-time search methods~\cite{zhai2025enhancing, lin2025qlass, xia2025agentrm}, which train separate reward or value models and perform multiple samplings at each step during inference.

Though our method is compatible with those test-time search techniques for LLM agents, for a more comprehensive view, we report in Table~\ref{tab:prm} some preliminary comparisons of SA\textsc{n}D with representative test-time search methods guided by PRMs~\cite{xia2025agentrm} and Q-value models~\cite{zhai2025enhancing, lin2025qlass}.
Note that the results are directly imported from the original papers and thus the base models might be slightly different.
We leave further integration of our agent tuning framework with advanced test-time search methods as future work.

\begin{table}[!t]
    \centering
    \resizebox{0.6\linewidth}{!}{
    \begin{tabular}{l | c}
    \toprule
    \textbf{Method} & \textbf{WebShop} \\
    \midrule
    Base & 55.3 \\
    SFT & 65.4 \\
    SAND (Iteration 1) & 68.5 \\
    SAND (Iteration 2) & 72.4 \\
    SAND (Iteration 3) & 71.8 \\
    \bottomrule
    \end{tabular}
    }
    \caption{Average rewards on WebShop.}
    \label{tab:webshop}
\end{table}

\section{Prompts}\label{app:prompt}

In this section, we provide the prompts used in our SA\textsc{n}D framework.
The prompt for execution-guided critique generation is shown in Figure \ref{fig:action_critique_prompt} and prompt for action deliberation synthesis is shown in Figure \ref{fig:action_deliberation_prompt}.
For evaluation on test set of ALFWorld and ScienceWorld, we follow the same prompts used in \citet{xiong2025mpo} for fair comparison, which is provided in Figure \ref{fig:alfworld_prompt} and Figure \ref{fig:sciworld_prompt}.

\newpage

\begin{figure*}[!t]
\definecolor{promptback}{RGB}{245,248,255}   
\definecolor{promptframe}{RGB}{124, 164, 207}    

\begin{tcblisting}{%
  breakable,
  colback=promptback,
  colframe=promptframe,
  arc=2pt,
  boxrule=0.5pt,
  leftrule=0.5pt,
  rightrule=0.5pt,
  toprule=0.5pt,
  bottomrule=0.5pt,
  title={Prompt for Execution-Guided Action Critique},
  listing only
}
### Background
{task_instruction}

### Current State
{interaction_history}

### Private Mental Simulations
You quietly imagined several futures that all start with the action **{sample_action}**.
Here is your simulated futures (keep it private):

{executed_rollout}

### Instructions
Write **one short paragraph (≤3 sentences)** titled exactly
`Action Evaluation:` that captures **your** intuitive judgement of
executing **{sampled_action}** now. In fluent prose, incorporate any of the following aspects as you see fit:

* Whether **{sampled_action}** in the current state is valid based on the environment feedback.
* Whether and how it might help advance the current progress toward important sub-goals or final goal of completing the task.
* Any task-relevant affordances or commonsense cues you should notice.  
* Frequent failure patterns or error loop you should be cautious for similar tasks.  
* A practical evaluation of the action **{sampled_action}** in the current state.

Do **not** directly quote or refer to the simulation log, and do **not** list items; blend them naturally into the paragraph.
Do **not** mention that the simulations exists or that you had outside help.

### Output Format
Action Evaluation: <your paragraph>
\end{tcblisting}

\caption{Prompt used for the execution-guided action critique.}
\label{fig:action_critique_prompt}
\end{figure*}

\begin{figure*}[!ht]
\definecolor{promptback}{RGB}{245,248,255}   
\definecolor{promptframe}{RGB}{124, 164, 207}    

\begin{tcblisting}{%
  breakable,
  colback=promptback,
  colframe=promptframe,
  arc=2pt,
  boxrule=0.5pt,
  leftrule=0.5pt,
  rightrule=0.5pt,
  toprule=0.5pt,
  bottomrule=0.5pt,
  title={Prompt for Action Deliberation Synthesis},
  listing only,
}
Background
{task_instrution}

### Current State
{interaction_history}

### Private Scratch-pad
You silently drafted several possible next actions with your intuitive judgement about each (these notes stay private):
- {candidate_action_1}: {critique_for_candidate_action_1}
- {candidate_action_2}: {critique_for_candidate_action_2}
- {candidate_action_3}: {critique_for_candidate_action_3}

### Very Important
Your final **Action** line must be **{expert_action}**. Everything you write has to lead naturally to this choice.

### Instructions
Generate reasoning thoughts following the instructions below:
Begin with a short one-sentence reflection of your previous action and your current situation. 
Then propose and list each candidate action from the scratch-pad with your own intuitive judgement, e.g., - <candidate action>: <your judgement>. 
Keep your judgement informative and avoid repeating generic evaluation statements.

Do **not** mention that the scratch-pad exists or that you got outside help.

### Output Format
Thought: <your one-sentence reflection>

- <candidate action>: <your judgement>
- <candidate action>: <your judgement>

<your comparison and rationale>
\end{tcblisting}

\caption{Prompt used for action deliberation synthesis.}
\label{fig:action_deliberation_prompt}
\end{figure*}

\begin{figure*}[!ht]
\definecolor{promptback}{RGB}{245,248,255}   
\definecolor{promptframe}{RGB}{124, 164, 207}    

\begin{tcblisting}{%
  breakable,
  colback=promptback,
  colframe=promptframe,
  arc=2pt,
  boxrule=0.5pt,
  leftrule=0.5pt,
  rightrule=0.5pt,
  toprule=0.5pt,
  bottomrule=0.5pt,
  title={Prompt for ALFWorld Tasks},
  listing only,
}
Interact with a household to solve a task. Imagine you are an intelligent agent in a household environment and your target is to perform actions to complete the task goal. At the beginning of your interactions, you will be given the detailed description of the current environment and your goal to accomplish. 
For each of your turn, you will be given the observation of the last turn. You should choose from two actions: "Thought" or "Action". If you choose "Thought", you should first think about the current condition and plan for your future actions, and then output your action in this turn. Your output must strictly follow this format:"Thought: your thoughts.\n Action: your next action"; If you choose "Action", you should directly output the action in this turn. Your output must strictly follow this format:"Action: your next action". 
The available actions are:
1. go to {recep}
2. take {obj} from {recep}
3. put {obj} in/on {recep}
4. open {recep}
5. close {recep}
6. toggle {obj} {recep}
7. clean {obj} with {recep}
8. heat {obj} with {recep}
9. cool {obj} with {recep}
where {obj} and {recep} correspond to objects and receptacles.
After your each turn, the environment will give you immediate feedback based on which you plan your next few steps. if the envrionment output "Nothing happened", that means the previous action is invalid and you should try more options.
Reminder: 
1. The action must be chosen from the given available actions. Any actions except provided available actions will be regarded as illegal.
2. Think when necessary, try to act directly more in the process.

Now, it's your turn and here is the task.
{task}
\end{tcblisting}

\caption{Prompt used for ALFWorld tasks.}
\label{fig:alfworld_prompt}
\end{figure*}

\begin{figure*}[!ht]
\definecolor{promptback}{RGB}{245,248,255}   
\definecolor{promptframe}{RGB}{124, 164, 207}    

\begin{tcblisting}{%
  breakable,
  colback=promptback,
  colframe=promptframe,
  arc=2pt,
  boxrule=0.5pt,
  leftrule=0.5pt,
  rightrule=0.5pt,
  toprule=0.5pt,
  bottomrule=0.5pt,
  title={Prompt for ScienceWorld Tasks},
  listing only,
}
You are a helpful assistant to do some scientific experiment in an environment.
In the environment, there are several rooms: kitchen, foundry, workshop, bathroom, outside, living room, bedroom, greenhouse, art studio, hallway
You should explore the environment and find the items you need to complete the experiment.
You can teleport to any room in one step.
All containers in the environment have already been opened, you can directly get items from the containers.
For each of your turn, you will be given the observation of the last turn. You should choose from two actions: "Thought" or "Action". If you choose "Thought", you should first think about the current condition and plan for your future actions, and then output your action in this turn. Your output must strictly follow this format:"Thought: your thoughts.\n Action: your next action"; If you choose "Action", you should directly output the action in this turn. Your output must strictly follow this format:"Action: your next action". Remember that you can only output one "Action:" in per response.

The available actions are:
open OBJ: open a container
close OBJ: close a container
activate OBJ: activate a device
deactivate OBJ: deactivate a device
connect OBJ to OBJ: connect electrical components
disconnect OBJ: disconnect electrical components
use OBJ [on OBJ]: use a device/item
look around: describe the current room
examine OBJ: describe an object in detail
look at OBJ: describe a container's contents
read OBJ: read a note or book
move OBJ to OBJ: move an object to a container
pick up OBJ: move an object to the inventory
pour OBJ into OBJ: pour a liquid into a container
mix OBJ: chemically mix a container
teleport to LOC: teleport to a specific room
focus on OBJ: signal intent on a task object
wait: task no action for 10 steps
wait1: task no action for a step

Now, it's your turn and here is the task.
{task}
\end{tcblisting}

\caption{Prompt used for ScienceWorld tasks.}
\label{fig:sciworld_prompt}
\end{figure*}

%% file: arxiv.bbl
\begin{thebibliography}{45}
\providecommand{\natexlab}[1]{#1}

\bibitem[{Achiam et~al.(2023)Achiam, Adler, Agarwal, Ahmad, Akkaya, Aleman, Almeida, Altenschmidt, Altman, Anadkat et~al.}]{achiam2023gpt}
Josh Achiam, Steven Adler, Sandhini Agarwal, Lama Ahmad, Ilge Akkaya, Florencia~Leoni Aleman, Diogo Almeida, Janko Altenschmidt, Sam Altman, Shyamal Anadkat, et~al. 2023.
\newblock Gpt-4 technical report.
\newblock \emph{arXiv preprint arXiv:2303.08774}.

\bibitem[{Chen et~al.(2023)Chen, Shu, Shareghi, Collier, Narasimhan, and Yao}]{chen2023fireact}
Baian Chen, Chang Shu, Ehsan Shareghi, Nigel Collier, Karthik Narasimhan, and Shunyu Yao. 2023.
\newblock Fireact: Toward language agent fine-tuning.
\newblock \emph{arXiv preprint arXiv:2310.05915}.

\bibitem[{Chen et~al.(2024)Chen, Liu, Wang, Zhang, Liu, Lin, Chen, and Zhao}]{chen2024agent}
Zehui Chen, Kuikun Liu, Qiuchen Wang, Wenwei Zhang, Jiangning Liu, Dahua Lin, Kai Chen, and Feng Zhao. 2024.
\newblock Agent-flan: Designing data and methods of effective agent tuning for large language models.
\newblock In \emph{Findings of the Association for Computational Linguistics ACL 2024}, pages 9354--9366.

\bibitem[{Chen et~al.(2025)Chen, Li, Huang, Du, Fang, and Zhou}]{chen2025atlas}
Zhixun Chen, Ming Li, Yuxuan Huang, Yali Du, Meng Fang, and Tianyi Zhou. 2025.
\newblock Atlas: Agent tuning via learning critical steps.
\newblock \emph{arXiv preprint arXiv:2503.02197}.

\bibitem[{Dubey et~al.(2024)Dubey, Jauhri, Pandey, Kadian, Al-Dahle, Letman, Mathur, Schelten, Yang, Fan et~al.}]{dubey2024llama}
Abhimanyu Dubey, Abhinav Jauhri, Abhinav Pandey, Abhishek Kadian, Ahmad Al-Dahle, Aiesha Letman, Akhil Mathur, Alan Schelten, Amy Yang, Angela Fan, et~al. 2024.
\newblock The llama 3 herd of models.
\newblock \emph{arXiv preprint arXiv:2407.21783}.

\bibitem[{Guan et~al.(2024)Guan, Joglekar, Wallace, Jain, Barak, Helyar, Dias, Vallone, Ren, Wei et~al.}]{guan2024deliberative}
Melody~Y Guan, Manas Joglekar, Eric Wallace, Saachi Jain, Boaz Barak, Alec Helyar, Rachel Dias, Andrea Vallone, Hongyu Ren, Jason Wei, et~al. 2024.
\newblock Deliberative alignment: Reasoning enables safer language models.
\newblock \emph{arXiv preprint arXiv:2412.16339}.

\bibitem[{Hu et~al.(2024)Hu, Wu, Zhu, Xianyu, Wang, Zhang, and Cao}]{hu2024openrlhf}
Jian Hu, Xibin Wu, Zilin Zhu, Xianyu, Weixun Wang, Dehao Zhang, and Yu~Cao. 2024.
\newblock Openrlhf: An easy-to-use, scalable and high-performance rlhf framework.
\newblock \emph{arXiv preprint arXiv:2405.11143}.

\bibitem[{Karanam et~al.(2024)Karanam, Jahanbakhsh, and Koyejo}]{karanam2024towards}
Arjun Karanam, Farnaz Jahanbakhsh, and Sanmi Koyejo. 2024.
\newblock Towards deliberating agents: Evaluating the ability of large language models to deliberate.
\newblock In \emph{NeurIPS 2024 Workshop on Behavioral Machine Learning}.

\bibitem[{Liang et~al.(2024)Liang, Song, Zheng, Wang, Yu, Li, Li, Wang, Wang, Xiong et~al.}]{liang2024internal}
Xun Liang, Shichao Song, Zifan Zheng, Hanyu Wang, Qingchen Yu, Xunkai Li, Rong-Hua Li, Yi~Wang, Zhonghao Wang, Feiyu Xiong, et~al. 2024.
\newblock Internal consistency and self-feedback in large language models: A survey.
\newblock \emph{arXiv preprint arXiv:2407.14507}.

\bibitem[{Lin et~al.(2025)Lin, Tang, Yao, Yin, Hu, Sun, and Chang}]{lin2025qlass}
Zongyu Lin, Yao Tang, Xingcheng Yao, Da~Yin, Ziniu Hu, Yizhou Sun, and Kai-Wei Chang. 2025.
\newblock Qlass: Boosting language agent inference via q-guided stepwise search.
\newblock \emph{arXiv preprint arXiv:2502.02584}.

\bibitem[{Madaan et~al.(2023)Madaan, Tandon, Gupta, Hallinan, Gao, Wiegreffe, Alon, Dziri, Prabhumoye, Yang et~al.}]{madaan2023self}
Aman Madaan, Niket Tandon, Prakhar Gupta, Skyler Hallinan, Luyu Gao, Sarah Wiegreffe, Uri Alon, Nouha Dziri, Shrimai Prabhumoye, Yiming Yang, et~al. 2023.
\newblock Self-refine: Iterative refinement with self-feedback.
\newblock \emph{Advances in Neural Information Processing Systems}, 36:46534--46594.

\bibitem[{Nakano et~al.(2021)Nakano, Hilton, Balaji, Wu, Ouyang, Kim, Hesse, Jain, Kosaraju, Saunders et~al.}]{nakano2021webgpt}
Reiichiro Nakano, Jacob Hilton, Suchir Balaji, Jeff Wu, Long Ouyang, Christina Kim, Christopher Hesse, Shantanu Jain, Vineet Kosaraju, William Saunders, et~al. 2021.
\newblock Webgpt: Browser-assisted question-answering with human feedback.
\newblock \emph{arXiv preprint arXiv:2112.09332}.

\bibitem[{Nguyen et~al.(2025)Nguyen, Chen, Wang, Wu, Park, Hu, Lyu, Wu, Aponte, Xia, Li, Shi, Chen, Lai, Xie, Kim, Zhang, Yu, Tanjim, Ahmed, Mathur, Yoon, Yao, Kveton, Kil, Nguyen, Bui, Zhou, Rossi, and Dernoncourt}]{nguyen-etal-2025-gui}
Dang Nguyen, Jian Chen, Yu~Wang, Gang Wu, Namyong Park, Zhengmian Hu, Hanjia Lyu, Junda Wu, Ryan Aponte, Yu~Xia, Xintong Li, Jing Shi, Hongjie Chen, Viet~Dac Lai, Zhouhang Xie, Sungchul Kim, Ruiyi Zhang, Tong Yu, Mehrab Tanjim, Nesreen~K. Ahmed, Puneet Mathur, Seunghyun Yoon, Lina Yao, Branislav Kveton, Jihyung Kil, Thien~Huu Nguyen, Trung Bui, Tianyi Zhou, Ryan~A. Rossi, and Franck Dernoncourt. 2025.
\newblock \href {https://doi.org/10.18653/v1/2025.findings-acl.1158} {{GUI} agents: A survey}.
\newblock In \emph{Findings of the Association for Computational Linguistics: ACL 2025}, pages 22522--22538, Vienna, Austria. Association for Computational Linguistics.

\bibitem[{Qiao et~al.(2024)Qiao, Fang, Zhang, Zhu, Chen, Deng, Jiang, Xie, Huang, and Chen}]{qiao2024agent}
Shuofei Qiao, Runnan Fang, Ningyu Zhang, Yuqi Zhu, Xiang Chen, Shumin Deng, Yong Jiang, Pengjun Xie, Fei Huang, and Huajun Chen. 2024.
\newblock Agent planning with world knowledge model.
\newblock \emph{Advances in Neural Information Processing Systems}, 37:114843--114871.

\bibitem[{Shi et~al.(2024)Shi, Yuan, Wu, Wang, and Feng}]{shi-etal-2024-direct}
Wentao Shi, Mengqi Yuan, Junkang Wu, Qifan Wang, and Fuli Feng. 2024.
\newblock \href {https://doi.org/10.18653/v1/2024.emnlp-main.138} {Direct multi-turn preference optimization for language agents}.
\newblock In \emph{Proceedings of the 2024 Conference on Empirical Methods in Natural Language Processing}, pages 2312--2324, Miami, Florida, USA. Association for Computational Linguistics.

\bibitem[{Shinn et~al.(2023)Shinn, Cassano, Gopinath, Narasimhan, and Yao}]{shinn2023reflexion}
Noah Shinn, Federico Cassano, Ashwin Gopinath, Karthik Narasimhan, and Shunyu Yao. 2023.
\newblock Reflexion: Language agents with verbal reinforcement learning.
\newblock \emph{Advances in Neural Information Processing Systems}, 36:8634--8652.

\bibitem[{Shridhar et~al.(2020)Shridhar, Yuan, C{\^o}t{\'e}, Bisk, Trischler, and Hausknecht}]{shridhar2020alfworld}
Mohit Shridhar, Xingdi Yuan, Marc-Alexandre C{\^o}t{\'e}, Yonatan Bisk, Adam Trischler, and Matthew Hausknecht. 2020.
\newblock Alfworld: Aligning text and embodied environments for interactive learning.
\newblock \emph{arXiv preprint arXiv:2010.03768}.

\bibitem[{Song et~al.(2024{\natexlab{a}})Song, Xiong, Zhao, Zhu, Wu, Wang, Li, Peng, and Li}]{song2024agentbank}
Yifan Song, Weimin Xiong, Xiutian Zhao, Dawei Zhu, Wenhao Wu, Ke~Wang, Cheng Li, Wei Peng, and Sujian Li. 2024{\natexlab{a}}.
\newblock Agentbank: Towards generalized llm agents via fine-tuning on 50000+ interaction trajectories.
\newblock In \emph{Findings of the Association for Computational Linguistics: EMNLP 2024}, pages 2124--2141.

\bibitem[{Song et~al.(2024{\natexlab{b}})Song, Yin, Yue, Huang, Li, and Lin}]{song-etal-2024-trial}
Yifan Song, Da~Yin, Xiang Yue, Jie Huang, Sujian Li, and Bill~Yuchen Lin. 2024{\natexlab{b}}.
\newblock \href {https://doi.org/10.18653/v1/2024.acl-long.409} {Trial and error: Exploration-based trajectory optimization of {LLM} agents}.
\newblock In \emph{Proceedings of the 62nd Annual Meeting of the Association for Computational Linguistics (Volume 1: Long Papers)}, pages 7584--7600, Bangkok, Thailand. Association for Computational Linguistics.

\bibitem[{Wang et~al.(2025{\natexlab{a}})Wang, Han, Zhang, Baldwin, and Li}]{wang-etal-2025-nat}
Renxi Wang, Xudong Han, Yixuan Zhang, Timothy Baldwin, and Haonan Li. 2025{\natexlab{a}}.
\newblock \href {https://aclanthology.org/2025.naacl-long.378/} {{NAT}: Enhancing agent tuning with negative samples}.
\newblock In \emph{Proceedings of the 2025 Conference of the Nations of the Americas Chapter of the Association for Computational Linguistics: Human Language Technologies (Volume 1: Long Papers)}, pages 7385--7398, Albuquerque, New Mexico. Association for Computational Linguistics.

\bibitem[{Wang et~al.(2022)Wang, Jansen, C{\^o}t{\'e}, and Ammanabrolu}]{wang2022scienceworld}
Ruoyao Wang, Peter Jansen, Marc-Alexandre C{\^o}t{\'e}, and Prithviraj Ammanabrolu. 2022.
\newblock Scienceworld: Is your agent smarter than a 5th grader?
\newblock \emph{arXiv preprint arXiv:2203.07540}.

\bibitem[{Wang et~al.(2025{\natexlab{b}})Wang, Wu, Xia, Yu, Rossi, McAuley, and Yao}]{wang2025dice}
Ruoyu Wang, Junda Wu, Yu~Xia, Tong Yu, Ryan~A Rossi, Julian McAuley, and Lina Yao. 2025{\natexlab{b}}.
\newblock Dice: Dynamic in-context example selection in llm agents via efficient knowledge transfer.
\newblock \emph{arXiv preprint arXiv:2507.23554}.

\bibitem[{Wang et~al.(2023)Wang, Wei, Schuurmans, Le, Chi, Narang, Chowdhery, and Zhou}]{wang2023selfconsistency}
Xuezhi Wang, Jason Wei, Dale Schuurmans, Quoc~V Le, Ed~H. Chi, Sharan Narang, Aakanksha Chowdhery, and Denny Zhou. 2023.
\newblock \href {https://openreview.net/forum?id=1PL1NIMMrw} {Self-consistency improves chain of thought reasoning in language models}.
\newblock In \emph{The Eleventh International Conference on Learning Representations}.

\bibitem[{Wei et~al.(2022)Wei, Wang, Schuurmans, Bosma, Xia, Chi, Le, Zhou et~al.}]{wei2022chain}
Jason Wei, Xuezhi Wang, Dale Schuurmans, Maarten Bosma, Fei Xia, Ed~Chi, Quoc~V Le, Denny Zhou, et~al. 2022.
\newblock Chain-of-thought prompting elicits reasoning in large language models.
\newblock \emph{Advances in neural information processing systems}, 35:24824--24837.

\bibitem[{Wu et~al.(2025)Wu, Xia, Yu, Chen, Harsha, Maharaj, Zhang, Bursztyn, Kim, Rossi, McAuley, Li, and Sinha}]{wu-etal-2025-doc}
Junda Wu, Yu~Xia, Tong Yu, Xiang Chen, Sai~Sree Harsha, Akash~V Maharaj, Ruiyi Zhang, Victor Bursztyn, Sungchul Kim, Ryan~A. Rossi, Julian McAuley, Yunyao Li, and Ritwik Sinha. 2025.
\newblock \href {https://doi.org/10.18653/v1/2025.acl-short.6} {Doc-react: Multi-page heterogeneous document question-answering}.
\newblock In \emph{Proceedings of the 63rd Annual Meeting of the Association for Computational Linguistics (Volume 2: Short Papers)}, pages 67--78, Vienna, Austria. Association for Computational Linguistics.

\bibitem[{Xia et~al.(2025{\natexlab{a}})Xia, Fan, Chen, Yan, Cong, Zhang, Lu, Lin, Liu, and Sun}]{xia2025agentrm}
Yu~Xia, Jingru Fan, Weize Chen, Siyu Yan, Xin Cong, Zhong Zhang, Yaxi Lu, Yankai Lin, Zhiyuan Liu, and Maosong Sun. 2025{\natexlab{a}}.
\newblock Agentrm: Enhancing agent generalization with reward modeling.
\newblock \emph{arXiv preprint arXiv:2502.18407}.

\bibitem[{Xia et~al.(2024{\natexlab{a}})Xia, Liu, Yu, Kim, Rossi, Rao, Mai, and Li}]{xia-etal-2024-hallucination}
Yu~Xia, Xu~Liu, Tong Yu, Sungchul Kim, Ryan Rossi, Anup Rao, Tung Mai, and Shuai Li. 2024{\natexlab{a}}.
\newblock \href {https://doi.org/10.18653/v1/2024.naacl-long.479} {Hallucination diversity-aware active learning for text summarization}.
\newblock In \emph{Proceedings of the 2024 Conference of the North American Chapter of the Association for Computational Linguistics: Human Language Technologies (Volume 1: Long Papers)}, pages 8665--8677, Mexico City, Mexico. Association for Computational Linguistics.

\bibitem[{Xia et~al.(2025{\natexlab{b}})Xia, Mukherjee, Xie, Wu, Li, Aponte, Lyu, Barrow, Chen, Dernoncourt, Kveton, Yu, Zhang, Gu, Ahmed, Wang, Chen, Deilamsalehy, Kim, Hu, Zhao, Lipka, Yoon, Huang, Wang, Mathur, Pal, Mukherjee, Zhang, Park, Nguyen, Luo, Rossi, and McAuley}]{xia-etal-2025-selection}
Yu~Xia, Subhojyoti Mukherjee, Zhouhang Xie, Junda Wu, Xintong Li, Ryan Aponte, Hanjia Lyu, Joe Barrow, Hongjie Chen, Franck Dernoncourt, Branislav Kveton, Tong Yu, Ruiyi Zhang, Jiuxiang Gu, Nesreen~K. Ahmed, Yu~Wang, Xiang Chen, Hanieh Deilamsalehy, Sungchul Kim, Zhengmian Hu, Yue Zhao, Nedim Lipka, Seunghyun Yoon, Ting-Hao~Kenneth Huang, Zichao Wang, Puneet Mathur, Soumyabrata Pal, Koyel Mukherjee, Zhehao Zhang, Namyong Park, Thien~Huu Nguyen, Jiebo Luo, Ryan~A. Rossi, and Julian McAuley. 2025{\natexlab{b}}.
\newblock \href {https://doi.org/10.18653/v1/2025.acl-long.708} {From selection to generation: A survey of {LLM}-based active learning}.
\newblock In \emph{Proceedings of the 63rd Annual Meeting of the Association for Computational Linguistics (Volume 1: Long Papers)}, pages 14552--14569, Vienna, Austria. Association for Computational Linguistics.

\bibitem[{Xia et~al.(2025{\natexlab{c}})Xia, Wang, Liu, Li, Yu, Chen, McAuley, and Li}]{xia-etal-2025-beyond}
Yu~Xia, Rui Wang, Xu~Liu, Mingyan Li, Tong Yu, Xiang Chen, Julian McAuley, and Shuai Li. 2025{\natexlab{c}}.
\newblock \href {https://aclanthology.org/2025.coling-main.719/} {Beyond chain-of-thought: A survey of chain-of-{X} paradigms for {LLM}s}.
\newblock In \emph{Proceedings of the 31st International Conference on Computational Linguistics}, pages 10795--10809, Abu Dhabi, UAE. Association for Computational Linguistics.

\bibitem[{Xia et~al.(2025{\natexlab{d}})Xia, Wu, Kim, Yu, Rossi, Wang, and McAuley}]{xia-etal-2025-knowledge}
Yu~Xia, Junda Wu, Sungchul Kim, Tong Yu, Ryan~A. Rossi, Haoliang Wang, and Julian McAuley. 2025{\natexlab{d}}.
\newblock \href {https://aclanthology.org/2025.naacl-long.216/} {Knowledge-aware query expansion with large language models for textual and relational retrieval}.
\newblock In \emph{Proceedings of the 2025 Conference of the Nations of the Americas Chapter of the Association for Computational Linguistics: Human Language Technologies (Volume 1: Long Papers)}, pages 4275--4286, Albuquerque, New Mexico. Association for Computational Linguistics.

\bibitem[{Xia et~al.(2024{\natexlab{b}})Xia, Yu, He, Zhao, McAuley, and Li}]{xia-etal-2024-aligning}
Yu~Xia, Tong Yu, Zhankui He, Handong Zhao, Julian McAuley, and Shuai Li. 2024{\natexlab{b}}.
\newblock \href {https://doi.org/10.18653/v1/2024.naacl-long.262} {Aligning as debiasing: Causality-aware alignment via reinforcement learning with interventional feedback}.
\newblock In \emph{Proceedings of the 2024 Conference of the North American Chapter of the Association for Computational Linguistics: Human Language Technologies (Volume 1: Long Papers)}, pages 4684--4695, Mexico City, Mexico. Association for Computational Linguistics.

\bibitem[{Xiong et~al.(2024{\natexlab{a}})Xiong, Payani, Yang, and Fekri}]{xiong2024deliberate}
Siheng Xiong, Ali Payani, Yuan Yang, and Faramarz Fekri. 2024{\natexlab{a}}.
\newblock Deliberate reasoning for llms as structure-aware planning with accurate world model.
\newblock \emph{arXiv preprint arXiv:2410.03136}.

\bibitem[{Xiong et~al.(2025)Xiong, Song, Dong, Zhao, Song, Wang, and Li}]{xiong2025mpo}
Weimin Xiong, Yifan Song, Qingxiu Dong, Bingchan Zhao, Feifan Song, Xun Wang, and Sujian Li. 2025.
\newblock Mpo: Boosting llm agents with meta plan optimization.
\newblock \emph{arXiv preprint arXiv:2503.02682}.

\bibitem[{Xiong et~al.(2024{\natexlab{b}})Xiong, Song, Zhao, Wu, Wang, Wang, Li, Peng, and Li}]{xiong-etal-2024-watch}
Weimin Xiong, Yifan Song, Xiutian Zhao, Wenhao Wu, Xun Wang, Ke~Wang, Cheng Li, Wei Peng, and Sujian Li. 2024{\natexlab{b}}.
\newblock \href {https://doi.org/10.18653/v1/2024.emnlp-main.93} {Watch every step! {LLM} agent learning via iterative step-level process refinement}.
\newblock In \emph{Proceedings of the 2024 Conference on Empirical Methods in Natural Language Processing}, pages 1556--1572, Miami, Florida, USA. Association for Computational Linguistics.

\bibitem[{Yang et~al.(2025)Yang, Li, Yang, Zhang, Hui, Zheng, Yu, Gao, Huang, Lv et~al.}]{yang2025qwen3}
An~Yang, Anfeng Li, Baosong Yang, Beichen Zhang, Binyuan Hui, Bo~Zheng, Bowen Yu, Chang Gao, Chengen Huang, Chenxu Lv, et~al. 2025.
\newblock Qwen3 technical report.
\newblock \emph{arXiv preprint arXiv:2505.09388}.

\bibitem[{Yao et~al.(2022)Yao, Chen, Yang, and Narasimhan}]{yao2022webshop}
Shunyu Yao, Howard Chen, John Yang, and Karthik Narasimhan. 2022.
\newblock Webshop: Towards scalable real-world web interaction with grounded language agents.
\newblock \emph{Advances in Neural Information Processing Systems}, 35:20744--20757.

\bibitem[{Yao et~al.(2023{\natexlab{a}})Yao, Yu, Zhao, Shafran, Griffiths, Cao, and Narasimhan}]{yao2023tree}
Shunyu Yao, Dian Yu, Jeffrey Zhao, Izhak Shafran, Tom Griffiths, Yuan Cao, and Karthik Narasimhan. 2023{\natexlab{a}}.
\newblock Tree of thoughts: Deliberate problem solving with large language models.
\newblock \emph{Advances in neural information processing systems}, 36:11809--11822.

\bibitem[{Yao et~al.(2023{\natexlab{b}})Yao, Zhao, Yu, Du, Shafran, Narasimhan, and Cao}]{yao2023react}
Shunyu Yao, Jeffrey Zhao, Dian Yu, Nan Du, Izhak Shafran, Karthik~R Narasimhan, and Yuan Cao. 2023{\natexlab{b}}.
\newblock \href {https://openreview.net/forum?id=WE_vluYUL-X} {React: Synergizing reasoning and acting in language models}.
\newblock In \emph{The Eleventh International Conference on Learning Representations}.

\bibitem[{Yuan et~al.(2025)Yuan, Chen, Xi, Ye, Du, and Chen}]{yuan2025agent}
Siyu Yuan, Zehui Chen, Zhiheng Xi, Junjie Ye, Zhengyin Du, and Jiecao Chen. 2025.
\newblock Agent-r: Training language model agents to reflect via iterative self-training.
\newblock \emph{arXiv preprint arXiv:2501.11425}.

\bibitem[{Yuan et~al.(2023)Yuan, Yuan, Li, Dong, Lu, Tan, Zhou, and Zhou}]{yuan2023scaling}
Zheng Yuan, Hongyi Yuan, Chengpeng Li, Guanting Dong, Keming Lu, Chuanqi Tan, Chang Zhou, and Jingren Zhou. 2023.
\newblock Scaling relationship on learning mathematical reasoning with large language models.
\newblock \emph{arXiv preprint arXiv:2308.01825}.

\bibitem[{Zelikman et~al.(2022)Zelikman, Wu, Mu, and Goodman}]{zelikman2022star}
Eric Zelikman, Yuhuai Wu, Jesse Mu, and Noah Goodman. 2022.
\newblock Star: Bootstrapping reasoning with reasoning.
\newblock \emph{Advances in Neural Information Processing Systems}, 35:15476--15488.

\bibitem[{Zeng et~al.(2024)Zeng, Liu, Lu, Wang, Liu, Dong, and Tang}]{zeng-etal-2024-agenttuning}
Aohan Zeng, Mingdao Liu, Rui Lu, Bowen Wang, Xiao Liu, Yuxiao Dong, and Jie Tang. 2024.
\newblock \href {https://doi.org/10.18653/v1/2024.findings-acl.181} {{A}gent{T}uning: Enabling generalized agent abilities for {LLM}s}.
\newblock In \emph{Findings of the Association for Computational Linguistics: ACL 2024}, pages 3053--3077, Bangkok, Thailand. Association for Computational Linguistics.

\bibitem[{Zhai et~al.(2025)Zhai, Yang, Xu, Feng, Yang, Ding, and Wang}]{zhai2025enhancing}
Yuanzhao Zhai, Tingkai Yang, Kele Xu, Dawei Feng, Cheng Yang, Bo~Ding, and Huaimin Wang. 2025.
\newblock Enhancing decision-making for llm agents via step-level q-value models.
\newblock In \emph{Proceedings of the AAAI Conference on Artificial Intelligence}, volume~39, pages 27161--27169.

\bibitem[{Zhang et~al.(2025)Zhang, Dai, Wu, Yang, Wang, Tang, and Liu}]{zhang2025survey}
Chen Zhang, Xinyi Dai, Yaxiong Wu, Qu~Yang, Yasheng Wang, Ruiming Tang, and Yong Liu. 2025.
\newblock A survey on multi-turn interaction capabilities of large language models.
\newblock \emph{arXiv preprint arXiv:2501.09959}.

\bibitem[{Zhu et~al.(2025)Zhu, Qiao, Ou, Deng, Lyu, Shen, Liang, Gu, Chen, and Zhang}]{zhu-etal-2025-knowagent}
Yuqi Zhu, Shuofei Qiao, Yixin Ou, Shumin Deng, Shiwei Lyu, Yue Shen, Lei Liang, Jinjie Gu, Huajun Chen, and Ningyu Zhang. 2025.
\newblock \href {https://aclanthology.org/2025.findings-naacl.205/} {{K}now{A}gent: Knowledge-augmented planning for {LLM}-based agents}.
\newblock In \emph{Findings of the Association for Computational Linguistics: NAACL 2025}, pages 3709--3732, Albuquerque, New Mexico. Association for Computational Linguistics.

\end{thebibliography}
